\documentclass[conference]{IEEEtran}
\usepackage{times}

\usepackage[numbers]{natbib}
\usepackage{multicol}
\usepackage[bookmarks=true]{hyperref}
\usepackage{algorithm}
\usepackage{algorithmic}
\usepackage{amsmath,amssymb}
\usepackage{graphicx}
\usepackage{stfloats}
\usepackage[dvipsnames]{xcolor}
\usepackage{caption}
\usepackage{lettrine}
\usepackage{booktabs}
\usepackage{makecell}
\usepackage{siunitx}
\usepackage{listings} 
\usepackage{xcolor}
\usepackage{multirow}
\usepackage{siunitx}
\usepackage[
  separate-uncertainty = true,
  multi-part-units = repeat
]{siunitx}
\begin{document}

\title{Jointly Learning Predicates and Actions Enables Zero-Shot Skill Composition}

\author{Benedict Quartey$^{1}$, Sebastian Castro$^{2}$, Eric Rosen$^{2}$, Wil Thomason$^{2}$, George Konidaris$^{1}$, Stefanie Tellex$^{1,2}$ \\ Brown University$^{1}$\hspace{1em}Robotics \& AI Institute$^{2}$
}

\newcommand{\approachname}{\textsc{Pacts}}
\newcommand{\stnote}[1]{\textcolor{blue}{\textbf{ST: #1}}}
\newcommand{\bennote}[1]{\textcolor{teal}{\textbf{BQ: #1}}}
\newcommand{\scnote}[1]{\textcolor{orange}{\textbf{SC: #1}}}
\newcommand{\ernote}[1]{\textcolor{red}{\textbf{ER: #1}}}

\twocolumn[{%
\renewcommand\twocolumn[1][]{#1}%
\maketitle
\begin{center}
\vspace*{-5mm}
    \centering
    \captionsetup{type=figure}
     \makebox[\textwidth][c]{\includegraphics[width=1\textwidth]{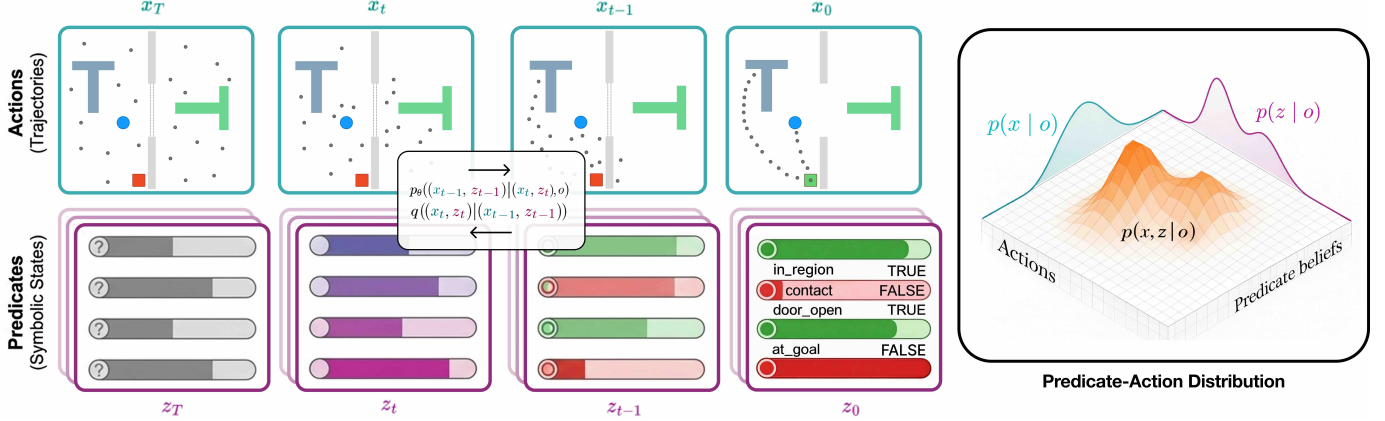}}%
    \captionof{figure}{\textbf{Predicate-Action Skills}. Conditioned on current observations $\mathbf{o}$, we model skills as a joint generative process over an action trajectory $\mathbf{x}$ and a predicate-belief trajectory $\mathbf{z}$ by learning the coupled distribution $p(\mathbf{x},\mathbf{z}\mid \mathbf{o})$. Starting from noise ($\mathbf{x}_T$,$\mathbf{z}_T$), our model iteratively refines both modalities to produce temporally coherent action–outcome rollouts $(\mathbf{x}_0,\mathbf{z}_0)$. 
    The resulting predicate-belief trajectory $\mathbf{z}_0$ provides an online symbolic interface for monitoring skill execution and planning-based skill composition using off-the-shelf planners.
    }
      \label{fig:splash}
\end{center}
}]

\begin{abstract}

Learning from Demonstration (LfD) enables robots to learn complex behaviors from expert examples, yet existing approaches often fail to generalize to new compositions of known skills without retraining. Modern generative policies model distributions over action trajectories alone, thus are unable to reason about the symbolic outcomes required for robust composition. We propose that skills should jointly model action trajectories and the symbolic outcomes they induce. To address this gap, we introduce Predicate-Action Skills (\approachname{}), a class of closed-loop visuomotor policies that model skills as a joint generative process over action and predicate belief trajectories, producing coherent action–outcome rollouts within a single model. Jointly generating actions and predicates enables \approachname{} to learn internal representations that improve both action generation and predicate classification. Furthermore, we demonstrate zero-shot composition of learned skills via planning by leveraging online predicate predictions from \approachname{} as a symbolic interface for sequencing and monitoring execution. 
Project website: \href{https://planpacts.github.io/}{planpacts.github.io}
\end{abstract}

\IEEEpeerreviewmaketitle

\section{Introduction}
\label{introduction}

Learning from Demonstration (LfD) has become a standard approach for teaching robots complex behaviors. Recent generative visuomotor policies are particularly compelling in this setting: by modeling distributions over action trajectories, they effectively capture multi-modal behavior. Yet, the gap between learning short-horizon behaviors and solving long-horizon, multi-step tasks remains open. In long-horizon execution, distribution shift compounds over time and the space of possible task variations and compositions grows combinatorially. Closing this gap typically requires large-scale demonstration datasets that cover many such variations or retraining whenever the desired composition changes. 

A promising route to long-horizon generalization is to introduce modular structure: reusable \emph{skills} that capture short-horizon behavior, and symbolic \emph{predicates} that describe state and skill effects in a form amenable to planning.
By abstracting both temporal behavior (via skills) and state (via predicates), prior work has shown that robots can generalize to unseen task compositions by sequencing learned skills to achieve goals \cite{konidaris2019necessityabstraction,garrett2020integratedtaskmotionplanning}.
However, these prior works typically \emph{decouple} these components, training one model to generate actions and a separate model to classify predicates from observations.
This separation obscures a central fact about manipulation: \emph{actions and symbolic outcomes are tightly coupled}.
For composition, what matters is not only which actions are plausible, but which symbolic conditions those actions will make true.

This paper starts from a simple alternative viewpoint: skills should not be represented only as a distribution over action trajectories, but as distributions over \emph{action--outcome rollouts}.
We represent outcomes as a \emph{predicate-belief trajectory}---a compact trace of soft truth values over predicates that describes what the model expects to become true (and with what confidence) as the skill unfolds.
Given an observation $\mathbf{o}$, we therefore model a coupled distribution over an action trajectory $\mathbf{x}$ and an outcome trace $\mathbf{z}$:
\[
(\mathbf{x},\mathbf{z}) \sim p(\mathbf{x},\mathbf{z}\mid \mathbf{o}).
\]
Sampling from this joint distribution yields coherent pairs, as the same sample that commits to an action mode also commits to the corresponding symbolic trace.
We instantiate this idea with \textbf{Predicate-Action Skills} (\approachname{}), closed-loop visuomotor policies that model skills as a \textbf{joint generative process} over action trajectories and predicate-belief trajectories.
\approachname{} iteratively refines both modalities to produce temporally aligned rollouts $(\mathbf{x}_0,\mathbf{z}_0)$ within a single model (see Fig.\ref{fig:splash}).
Crucially, the resulting predicate-belief trajectory $\mathbf{z}_0$ provides a practical interface for planning-based skill composition: it can be used to check preconditions and effects, monitor execution, and trigger replanning when expectations are violated---enabling zero-shot composition of learned skills with off-the-shelf planners. \looseness=-1

Our formulation supports both diffusion-style denoising and flow-matching objectives, and instantiates naturally with different sequence backbones (e.g., UNet- and Transformer-based policies). We systematically evaluate \approachname{} on a compositional 2D benchmark for comprehensive ablations and on two 3D RoboMimic tasks~\cite{robomimic2021,mandlekar2023mimicgen}, spanning a broad grid of model variants. Across these settings, jointly generating actions and predicate beliefs maintains competitive performance and often improves both action performance and predicate prediction relative to action-only generative policies and pipelines that pair action generation with a separate predicate predictor. Finally, we demonstrate planning-based composition of learned subskills in simulation and in a real-world environment. 

In summary, we make three contributions: \textbf{(1)} a joint generative skill formulation that models action trajectories and predicate traces to produce coherent action--outcome rollouts; \textbf{(2)} planning-based skill composition using predicate-belief traces as an online interface for sequencing and monitoring with off-the-shelf planners; and \textbf{(3)} an open-source skill segmentation and labeling toolkit that converts monolithic demonstrations into skill-centric datasets for joint $(\mathbf{x},\mathbf{z})$ learning in real-world deployments. \looseness=-1

\section{Related Work}
\label{sec:related_work}

\paragraph{Generative visuomotor policies for imitation learning}
Behavior cloning learns skills by mapping observations to actions~\cite{torabi2018behavioralcloningobservation}, and recent work increasingly models \emph{distributions} over action trajectories to capture multi-modality and long-horizon structure. Diffusion-based policies~\cite{chi2024diffusionpolicy} and conditional flow matching~\cite{chisari2024flowmatching} are prominent instantiations, with complementary approaches improving data efficiency via pretrained vision-language representations~\cite{zhang2024vlms,black2024pi0} or architectural/representational structure (e.g., equivariance and object-/affordance-centric models)~\cite{wang2024equivariantdiffusionpolicy,yuan2022sornet,rana2024affordancecentricpolicylearningsample}. Diffusion has also been used for planning: Diffuser~\cite{janner2022planning} denoises state--action trajectories to enable flexible test-time conditioning, whereas Diffusion Policy~\cite{chi2024diffusionpolicy} treats control as learning the distribution over action sequences, conditioning on observations without denoising them for efficient real-time inference.
We adopt this conditional-policy perspective, but depart from action-only modeling by learning a coupled generator $p(\mathbf{x},\mathbf{z}\mid\mathbf{o})$ that jointly samples actions and a low-dimensional predicate-belief rollout $\mathbf{z}$ aligned with PDDL preconditions/effects, yielding an outcome-aware interface for execution-time monitoring and planning-based composition.
Vision--language--action models similarly couple control with language semantics, but their semantics are typically part of the conditioning interface rather than an explicit planner-consumable outcome trajectory.

\paragraph{Skills, abstraction, and planning-based composition}
Automated task planning in robotics \cite{ghallab2016automated} specifies a domain (often in PDDL \cite{ghallab1998pddl}) in terms of abstract actions with preconditions and effects, and uses search to synthesize sequences that achieve goal conditions.
In robotic systems, these abstract actions are typically instantiated as hand-designed controllers (e.g., finite-state machines \cite{harel1987statecharts} or behavior trees \cite{colledanchise2018bt}), and predicates are grounded through engineered perception and state estimation modules \cite{migimatsu2022symbolicstateestimationpredicates}. Several frameworks support such symbolic planning architectures \cite{diab2020skillman,martin2021plansys2,mayr2023skiros2,paulius2025bootstrapping, quartey2025verifiably}. 

Recent learning-based approaches aim to reduce manual engineering by learning both skill policies and predicate evaluators from data: policies are learned via behavior cloning or reinforcement learning, while predicates are grounded either with supervised classifiers or with pretrained vision--language models that infer symbolic propositions from observations. A common design pattern in learned planning pipelines is to train a policy separately from a predicate evaluator (classifier or pretrained model) and use the latter as the planning/monitoring interface. However, in long-horizon manipulation, actions and symbolic outcomes are tightly coupled: different action \emph{modes} correspond to different predicate transitions, and treating them as separate predictions can lead to inference-time inconsistency. \approachname{} addresses this coupling directly by modeling a skill as a \emph{joint} generative process over action trajectories and predicate-belief trajectories. By sampling $(\mathbf{x},\mathbf{z})$ from a single model, the predicted outcome trace remains aligned with the sampled action mode. \looseness=-1

\section{Jointly Modeling Predicates and Actions}
\label{sec:method} 

\begin{figure*}   
  \centering
  \includegraphics[width=1\textwidth]{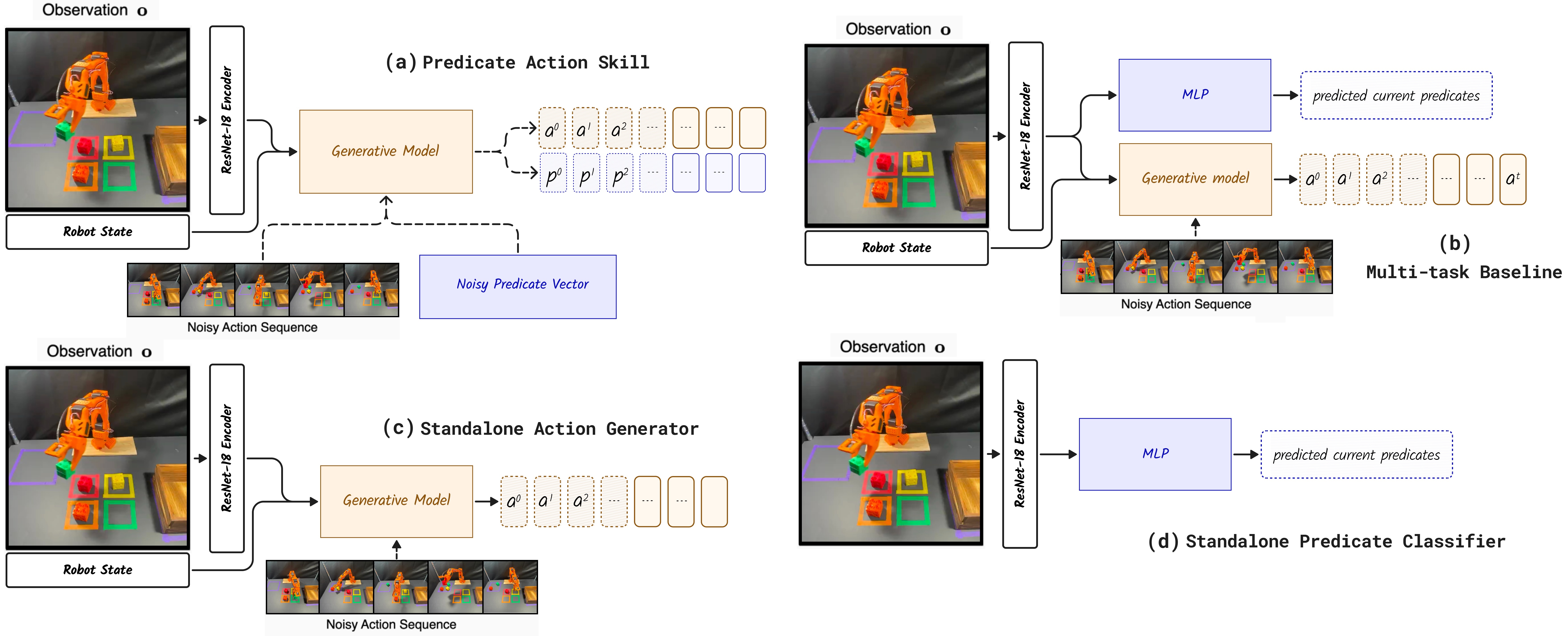}
\caption{\textbf{Architectures and baselines.}
\textbf{(a)} \approachname{} uses a \emph{single} conditional generative model to sample \emph{joint} action--predicate rollouts, refined together via DDPM denoising or flow matching.
To \emph{disentangle} joint modeling from multi-task learning, we include a multi-task baseline \textbf{(b)} that shares the same observation encoder but predicts predicates and actions with \emph{separate} heads trained with a combined loss. Standalone baselines isolate each component: \textbf{(c)} a generative \emph{action-only} generator that generates actions, and \textbf{(d)} a \emph{predicate-only} model predicts predicates from observations.}
\label{fig:architectures}
  \vspace*{-3mm}
\end{figure*}

\subsection{Problem definition: joint action--outcome skill model}

We model a closed-loop skill as a conditional distribution over a horizon-$H$ action trajectory
$\mathbf{x}=(\mathbf{a}_t,\dots,\mathbf{a}_{t+H-1})\in\mathbb{R}^{H\times d_a}$, where each $\mathbf{a}_{t+h}\in\mathbb{R}^{d_a}$ is a single control action. To support planning-based skill composition, we additionally model the symbolic outcomes induced by actions using a
\emph{predicate-belief trajectory}
$\mathbf{z}=(\mathbf{z}_{t+1},\dots,\mathbf{z}_{t+H})\in[0,1]^{H\times J}$,
where $\mathbf{z}_{t+h,j}$ is the model's confidence that predicate $j\in\{1,\dots,J\}$ will hold at future step $t+h$. 

Given the current observation $\mathbf{o}_t$, \approachname{} learn the coupled distribution $p_\theta(\mathbf{x},\mathbf{z}\mid \mathbf{o}_t)$ so that a single sample yields a coherent action--outcome rollout. We implement this by defining the joint trajectory variable $\mathbf{y} \triangleq [\mathbf{x}\,|\,\mathbf{z}] \in \mathbb{R}^{H\times(d_a+J)}$ and learning a single conditional generative model $p_\theta(\mathbf{y}\mid \mathbf{o}_t)$. This joint modeling choice is the key distinction from decoupled or multi-task baselines: $\mathbf{z}$ is not produced by an independent classifier head, but is part of the generated trajectory sample.

\paragraph{Predicate-belief representation.}
In the dataset, predicates are provided as binary labels in $\{0,1\}^J$.
We treat predicate channels as continuous during generation by linearly normalizing labels to $[-1,1]$ and learning to generate them jointly with actions. At inference, we map the generated predicate channels back to $[0,1]$, yielding soft beliefs that can be thresholded when discrete predicate evaluations are required (e.g., checking preconditions/effects during planning).

\subsection{DDPM formulation}
\label{sec:ddpm_formulation}
Following diffusion policy \cite{chi2024diffusionpolicy}, we model the conditional trajectory distribution $p(\mathbf{y}\mid \mathbf{o}_t)$:
we condition on $\mathbf{o}_t$ but do not denoise observations or infer future observation (or state), avoiding the cost of modeling observations and trajectories \cite{janner2022planning}. We apply denoising diffusion probabilistic modeling (DDPM) ~\cite{ho2020denoising} to the joint variable $\mathbf{y}=[\mathbf{x}\,|\,\mathbf{z}]$. Let $q(\mathbf{y}^k\mid \mathbf{y}^{k-1})$ denote the forward noising process with a standard noise schedule
$\{\alpha_k\}_{k=1}^{K}$. We train a noise-prediction network $\epsilon_\theta$, conditioned on $\mathbf{o}_t$, to predict the injected noise:
\begin{equation}
\mathcal{L}_{\textsc{ddpm}}
=
\mathbb{E}_{\mathbf{o}_t,\mathbf{y}^0,k,\epsilon}
\left[
\left\|
\epsilon_\theta(\mathbf{o}_t,\mathbf{y}^k,k)-\epsilon
\right\|^2
\right],
\end{equation}
where $\mathbf{y}^k$ is obtained by noising the clean joint trajectory $\mathbf{y}^0=[\mathbf{x}\,|\,\mathbf{z}]$ at \emph{denoising step} $k$.
At inference, we sample $\mathbf{y}^K\sim\mathcal{N}(0,I)$ and iteratively denoise to obtain $\mathbf{y}^0$.

\subsection{Conditional Flow Matching formulation}
\label{sec:flow_matching}
The joint object $\mathbf{y}=[\mathbf{x}\,|\,\mathbf{z}]$ can also be learned via \emph{conditional flow matching} (CFM)~\cite{tong2023improving,lipman2022flow,black2024pi0}, which targets the same conditional distribution $p_\theta(\mathbf{y}\mid \mathbf{o}_t)$ but uses a different objective and sampler.
Let $\mathbf{y}^1\sim\mathcal{N}(0,I)$ and define the optimal transport path between data $\mathbf{y}^0$ and noise $\mathbf{y}^1$ as
\begin{equation}
\mathbf{y}_\tau = \tau\,\mathbf{y}^1 + (1-\tau)\,\mathbf{y}^0,\qquad \tau\in[0,1],
\end{equation}
with constant target velocity $\mathbf{u}_\tau=\frac{d}{d\tau}\mathbf{y}_\tau=\mathbf{y}^1-\mathbf{y}^0$.
A conditional vector field $v_\theta(\mathbf{o}_t,\mathbf{y}_\tau,\tau)$ is trained by regression:
\begin{equation}
\mathcal{L}_{\textsc{cfm}}
=
\mathbb{E}_{\mathbf{o}_t,\mathbf{y}^0,\mathbf{y}^1,\tau}
\left[
\left\|
v_\theta(\mathbf{o}_t,\mathbf{y}_\tau,\tau) - (\mathbf{y}^1-\mathbf{y}^0)
\right\|^2
\right].
\end{equation}
At inference, samples are obtained by integrating from noise ($\tau{=}1$) to data ($\tau{=}0$), yielding $\mathbf{y}^0$.

Importantly, DDPM and flow matching instantiate the same modeling choice---learning the coupled action--outcome distribution
$p_\theta(\mathbf{x},\mathbf{z}\mid\mathbf{o}_t)$---but provide complementary training lenses with different sampling trade-offs.

\subsection{Architecture}
\label{sec:architectures}
As shown in Fig.~\ref{fig:architectures}(a), \approachname{} factorizes into (i) an observation encoder and (ii) a conditional trajectory generator. The observation encoder maps the current observation $\mathbf{o}_t$ into a latent embedding used to condition the generator. The generator implements either a DDPM noise-prediction network $\epsilon_\theta(\mathbf{o}_t,\mathbf{y}^k,k)$ (Sec.~\ref{sec:ddpm_formulation}) or a CFM velocity field $v_\theta(\mathbf{o}_t,\mathbf{y}_\tau,\tau)$ (Sec.~\ref{sec:flow_matching}), operating on the joint trajectory $\mathbf{y}=[\mathbf{x}\,|\,\mathbf{z}]$.

We instantiate the trajectory generator with two backbones:
(i) a temporal UNet that applies 1D convolutions over the horizon dimension, and
(ii) a Transformer that treats time steps as tokens and models long-range temporal dependencies via self-attention.
In both cases, conditioning enters through (a) the observation embedding and (b) the diffusion/flow time parameter (discrete $k$ for DDPM, continuous $\tau$ for CFM), injected via standard conditional modulation.

Like prior trajectory-generating visuomotor policies, \approachname{} can be executed in a receding-horizon manner: at each control cycle, the policy conditions on the current observation $\mathbf{o}_t$, samples a length-$H$ action chunk $\mathbf{x}$ (and its paired predicate-belief trajectory $\mathbf{z}$), executes a short prefix of $\mathbf{x}$, then re-observes and re-samples.

\begin{algorithm}[t]
\caption{Planning with Predicted Predicate Beliefs}
\label{alg:predicate_belief_planning}
\begin{algorithmic}[1]
\REQUIRE Skill library $\mathcal{S}$ with \approachname{} $\{p_\theta(\mathbf{x},\mathbf{z}\mid \mathbf{o}, s)\}_{s\in\mathcal{S}}$; planning domain $\mathcal{D}$; PDDL problem instance $\mathcal{P}$ with initial predicate state $\hat{\mathbf{p}}_0$ and goal $g$
\STATE $t \leftarrow 0$; \ $\hat{\mathbf{p}}_t \leftarrow \hat{\mathbf{p}}_0$; \ observe $\mathbf{o}_t$
\WHILE{goal $g$ not satisfied under $\hat{\mathbf{p}}_t$}
    \STATE $\pi \leftarrow \textsc{Plan}(\mathcal{D}, \hat{\mathbf{p}}_t, g)$ \COMMENT{symbolic plan in predicate space}
    \STATE $s \leftarrow \pi[0]$ \COMMENT{next planned skill}
    \WHILE{\textbf{not} \textsc{EffectsSatisfied}$(s,\hat{\mathbf{p}}_t)$ \textbf{and} \textbf{not} \textsc{TerminateSkill}$(s)$}
        \STATE Sample rollout $(\mathbf{x},\mathbf{z}) \sim p_\theta(\mathbf{x},\mathbf{z}\mid \mathbf{o}_t,s)$
        \STATE Execute first action (or short prefix) from $\mathbf{x}$ and observe $\mathbf{o}_{t+1}$
        \STATE $\hat{\mathbf{p}}_{t+1} \leftarrow \textsc{Thresh}(\mathbf{z}_{t+1})$ \COMMENT{predicted next predicate state}
        \STATE $t \leftarrow t+1$; \ $\hat{\mathbf{p}}_t \leftarrow \hat{\mathbf{p}}_{t+1}$; \ $\mathbf{o}_t \leftarrow \mathbf{o}_{t+1}$
    \ENDWHILE
\ENDWHILE
\end{algorithmic}
\end{algorithm}

\section{Planning-based Skill Composition with Predicate Beliefs}
\label{sec:planning}
A key benefit of modeling skills as joint action--outcome rollouts is that the generated predicate-belief trajectory provides a symbolic interface for composition. This interface is compatible with formal robot planning pipelines that represent tasks using symbolic or temporal-logic structure~\cite{garrett2020integratedtaskmotionplanning, paulius2025bootstrapping, quartey2025verifiably}. We assume a PDDL symbolic planning domain specifying skill-level operators with preconditions and effects over predicates, and a PDDL problem instance specifying an initial predicate state and a goal. Given a library of learned skills, an off-the-shelf planner synthesizes a sequence of skills to achieve a goal.
During execution, \approachname{} produces online predicate-belief estimates that serve as the planner state, enabling the system to (i) check whether a skill's intended effects are satisfied and (ii) replan from an updated symbolic state.

\paragraph{Predicate beliefs as an execution interface}
At each decision point, the robot observes $\mathbf{o}_t$ and (for the selected skill $s$) samples a joint rollout $(\mathbf{x},\mathbf{z})\sim p_\theta(\mathbf{x},\mathbf{z}\mid \mathbf{o}_t,s)$.
The belief trajectory $\mathbf{z}\in[0,1]^{H\times J}$ provides a compact prediction of how symbolic conditions are expected to evolve as the skill unfolds.
We obtain a discrete predicate valuation by thresholding beliefs (e.g., $\mathbf{z}_{t+h,j}>0.5$), and use the predicted next-step predicates $\hat{\mathbf{p}}_{t+1}=\textsc{Thresh}(\mathbf{z}_{t+1})$ as the symbolic state for effect checking and replanning.
Although $\hat{\mathbf{p}}_{t+1}$ is obtained from the model's prediction $\mathbf{z}_{t+1}$ (made when sampling at time $t$, before observing $\mathbf{o}_{t+1}$), beliefs are re-grounded at every step because the next rollout is conditioned on the latest observation $\mathbf{o}_{t+1}$.

\paragraph{Plan--execute--check--replan}
Algorithm~\ref{alg:predicate_belief_planning} summarizes our composition loop.
We replan at a regular cadence (after each executed skill), using the updated predicted predicate state.
This yields a simple integration with off-the-shelf planners: the planner proposes a next skill based on the current symbolic state, and execution proceeds step-by-step until the skill's effects are satisfied (or a termination condition is met), at which point we replan.

\subsection{Skill segmentation and labeling toolkit.}
To make real-world deployment tractable, we introduce a skill segmentation and labeling toolkit that converts monolithic demonstrations into training-ready, skill-centric datasets for joint $(\mathbf{x},\mathbf{z})$ learning.
The toolkit operates natively on LeRobot \cite{cadene2024lerobot} episodic datasets (state/action logs with video observations) and provides three core capabilities:
(i) sparse keyframe annotation of grounded predicates defined directly from the PDDL domain/problem files,
(ii) automatic conversion of sparse annotations into dense per-timestep predicate labels via forward-fill interpolation, and
(iii) skill-level segmentation by identifying goal-achieving predicate configurations and backward-filling executing phases to extract contiguous per-skill segments).

\section{Evaluation}

\begin{figure}
\centering
  \includegraphics[scale=0.125]{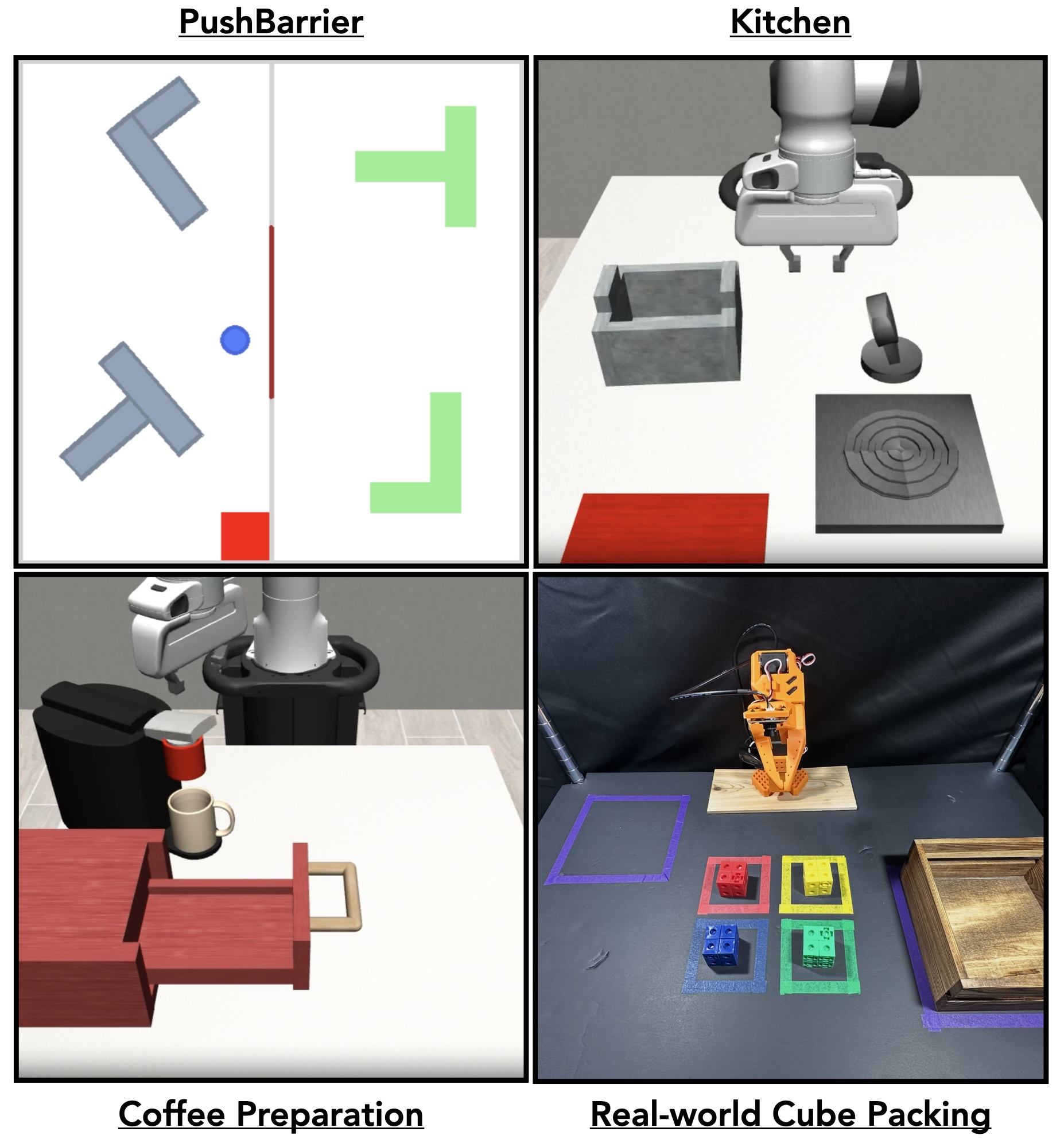}
\caption{\textbf{Evaluation environments.}
We evaluate \approachname{} across (a) \textbf{PushBarrier}, a controlled 2D compositional environment for systematic ablations and planner-based skill sequencing; (b) \textbf{Kitchen} and (c) \textbf{Coffee Preparation} from RoboMimic/MimicGen, which test joint action--predicate modeling under realistic 3D visual manipulation with simulator predicate oracles; and (d) a \textbf{real-world cube packing} task, where a robot packs four colored cubes into a container.}
    \vspace*{-4mm}
  \label{fig:environment}
\end{figure} 

We evaluate \approachname{} across three(3) tasks in simulation and a real-world multi-step task. Our experiments are designed to answer three (3) key questions:
\begin{enumerate}
    \item How does joint predicate--action generation affect action performance and predicate classification accuracy?
    \item How do modeling choices--DDPM denoising versus conditional flow matching, and UNet versus Transformer backbones--impact performance?
    \item Can the learned predicate-belief interface support planning-based recomposition and replanning-based recovery on novel goals without retraining?
\end{enumerate}

\subsection{Tasks and datasets}
\label{sec:tasks_datasets}

We evaluate \approachname{} in three settings that trade off control, realism, and compositional complexity (see Fig. \ref{fig:environment}).

\vspace{2mm}

\noindent\textbf{2D PushBarrier (compositional benchmark).}
We introduce a multi-step 2D manipulation task adapted from the popular Push-T environment \cite{chi2024diffusionpolicy}.
The workspace contains two objects (a \texttt{T} block and an \texttt{L} block) that must each be pushed from their respective start regions to corresponding goal regions.
Successful completion requires not only reaching the goal regions but also achieving the correct final \emph{orientation/alignment} of each object with its goal marker.
A barrier separates the start and goal regions, and the robot must first press a button to open the barrier before either object can be transferred to the goal side.
This environment provides a controlled setting with clear compositional structure and is our primary benchmark for systematic evaluations and ablations.

We use PushBarrier in two complementary evaluation modes. 
To address Questions (1) and (2) in an end-to-end setting, we train \emph{single} long-horizon policies on the full task and compare task performance and predicate classification accuracy against our baselines.
To address Questions (3) in a compositional setting, we construct a library of short-horizon skills and a task-relevant predicate set, segment demonstrations into per-skill datasets, and train a separate \approachname{} model for each skill. We then evaluate planning-based compositions that require sequencing skills, including \emph{novel skill sequences} not observed during training.

\vspace{2mm}

\noindent\textbf{RoboMimic (3D manipulation).}
To evaluate under realistic visual and geometric complexity, we additionally evaluate on two RoboMimic \cite{robomimic2021} tasks with MimicGen \cite{mandlekar2023mimicgen} demonstration data:
\textbf{Kitchen} and \textbf{Coffee Preparation}.
These tasks involve long-horizon manipulation under visually grounded observations and serve as a complementary test of whether joint modeling improves or maintains performance beyond the controlled 2D benchmark.
For each task, we define a predicate set aligned with task structure and implement predicate oracles inside the simulator that compute ground-truth predicate values from simulator state.
These predicate labels are used for both training and evaluation, enabling direct comparisons of predicate classification accuracy across methods. We use the base-difficulty (\texttt{d0}) variants, converted to absolute action space, with 200 demonstrations per task: \texttt{kitchen\_d0} and \texttt{coffee\_preparation\_d0}.

\vspace{2mm}

\noindent\textbf{Real-world demonstration}
We include a real-world on a cube packing task. A tabletop robot must pick and place four colored cubes from fixed start regions into a container. We collect \emph{monolithic}, long-horizon demonstrations that complete the full task (packing all four cubes), then use our segmentation and labeling toolkit to decompose each episode into skill-centric sub-demonstrations corresponding to per-cube packing behaviors (e.g., \texttt{pack\_red}, \texttt{pack\_green}, etc.). We train a separate \approachname{} model for each subskill and compose them with an off-the-shelf PDDL planner. This enables recompositions beyond the original demonstrations, such as executing individual packing skills in isolation or packing only a specified subset of cubes (e.g., ``pack red and yellow''), providing a concrete test of zero-shot compositionality in the real world.

\subsection{Baselines and ablations}
\label{sec:baselines}

We design strong baselines to isolate the benefit of modeling the joint action--outcome distribution $p_\theta(\mathbf{x},\mathbf{z}\mid\mathbf{o}_t)$ and sampling paired action--predicate rollouts in \approachname{}. We begin with baselines that model each conditional in isolation (action-only and predicate-only). We then include a shared-encoder multi-task baseline that predicts actions and predicates with separate heads trained end-to-end, which isolates the effect of \emph{joint sampling} from standard multi-task representation sharing.

\paragraph{Action-only generative policy (Fig.~\ref{fig:architectures}(c)).}
A diffusion/flow policy that models only $p(\mathbf{x}\mid\mathbf{o}_t)$ by generating an action trajectory $\mathbf{x}$ without predicate channels. This represents the standard action-only generative-policy baseline and tests whether adding outcome modeling degrades (or improves) action quality.

\paragraph{Standalone predicate predictor (Fig.~\ref{fig:architectures}(d)).}
A predicate-only model that predicts predicate beliefs from observations, modeling $p(\mathbf{z}\mid\mathbf{o}_t)$. This quantifies predicate prediction performance in isolation and mirrors common pipelines where predicates are treated purely as a discriminative output.

\paragraph{Multi-task baseline (Fig.~\ref{fig:architectures}(b)).}
To disentangle joint modeling from multi-task learning, we include a baseline with a shared observation encoder and two output heads trained end-to-end with a combined loss.
This corresponds to learning a factorized conditional model
\begin{equation}
p(\mathbf{x},\mathbf{z}\mid \mathbf{o}_t) \approx p(\mathbf{x}\mid \mathbf{o}_t)\,p(\mathbf{z}\mid \mathbf{o}_t),
\end{equation}
where the action head is a conditional diffusion/flow policy trained to model $p(\mathbf{x}\mid\mathbf{o}_t)$ (using the same DDPM/CFM objectives as Sec.~\ref{sec:ddpm_formulation}--\ref{sec:flow_matching}),
and the predicate head is a discriminative predictor trained with binary cross-entropy on ground-truth predicates.
The combined objective is
\begin{equation}
\mathcal{L}_{\text{total}}=\mathcal{L}_{\text{action}}+\mathcal{L}_{\text{predicate}}.
\end{equation}
Although the encoder is shared, the two heads produce outputs independently at inference, so this baseline cannot encourage action--predicate consistency at sampling time.

\vspace{2mm}

\noindent\textbf{Objective and backbone sweeps.}
All methods (\approachname{} and baselines) are evaluated across both training objectives (DDPM and CFM) and both backbone choices (UNet and Transformer), so differences can be attributed to joint modeling rather than a particular generative objective or architecture.

\subsection{Metrics}
\label{sec:metrics}

We report the following metrics.
\paragraph{Policy evaluation}
We evaluate methods by rolling out policies in the environment and reporting task performance.
In PushBarrier we report \emph{Max Reward} ($\uparrow$) and \emph{Steps to Max Reward} ($\downarrow$). In RoboMimic, we follow the benchmark protocol and report rollout success/reward statistics. Results are mean $\pm$ standard error over seven (7) seeded rollouts.

\paragraph{Predicate classification evaluation}
We compare predicted predicates to ground-truth oracle labels and report macro \emph{F1}, \emph{recall}, \emph{precision}, and \emph{accuracy}.  For belief outputs $\mathbf{z}\in[0,1]^{H\times J}$, we threshold at $0.5$ to obtain $\hat{\mathbf{p}}\in\{0,1\}^J$.  Predicate metrics are \emph{macro-averaged across predicates} so that each predicate contributes equally, preventing frequent predicates from dominating the score.

\paragraph{Predicate--action coherence}
To test whether predicted outcomes match the outcomes induced by executed actions, we compute horizon-indexed coherence curves over open-loop action chunks.
At each prediction cycle $\tau$, the policy predicts $\hat{\mathbf{z}}_{\tau+1:\tau+H}$ and executes its generated action chunk; we then compare $\hat{\mathbf{z}}_{\tau+h}$ to oracle predicates $\mathbf{p}_{\tau+h}$ for $h\in\{1,\dots,H\}$.
We report \textsc{CrossCons}$(h)$ (thresholded agreement, $\uparrow$), \textsc{CrossBrier}$(h)$ (MSE on beliefs, $\downarrow$), and \textsc{CrossNLL}$(h)$ (BCE on beliefs, $\downarrow$), along with majority/base-rate reference baselines.\looseness=-1

\begin{table*}[t]
\centering
\caption{\textbf{Single-policy rollout evaluation on PushBarrier.}
We compare \approachname{} to action-only, predicate-only, and shared-encoder multi-task baselines when learning a \emph{single} policy for the full task. This isolates whether joint action--outcome modeling degrades action performance and whether it improves predicate prediction in the absence of any skill segmentation or planning. Policy metrics are mean$\pm$standard error over seven (7) seeded rollouts}
\label{tab:rollout-eval-pushtlbarrier_blocked}
\resizebox{\textwidth}{!}{
\begin{tabular}{@{}llcccccc@{}}
\toprule
\textbf{Setting} & \textbf{Approach} &
\multicolumn{1}{c}{\makecell{Policy Evaluation}} &
\multicolumn{5}{c}{\makecell{Predicate Classification Evaluation}} \\
\cmidrule(lr){3-3} \cmidrule(lr){4-8}
(Objective + Backbone)& &
Max Reward $\uparrow$ &
F1 $\uparrow$ & Recall $\uparrow$ & Precision $\uparrow$ & Accuracy $\uparrow$ & Observations \\
\midrule
\midrule

\multirow{3}{*}{\makecell[l]{DDPM + UNet}}
&Action-Only Generator + Predicate-Only Predictor & 0.83 $\pm$ 0.09 & \SI{0.9476}{} & \SI{0.9547}{} & \SI{0.9471}{} & \SI{79.40}{\%} & 2403 \\
& Multi-task Baseline (Shared Encoder)  & 0.77 $\pm$ 0.13 & \SI{0.9764}{} & \SI{0.9712}{} & \SI{0.9834}{} & \SI{89.39}{\%} & 2705 \\
& \textbf{Predicate Action Skill (\approachname{})}            & \textbf{1.00 $\pm$ 0.00} & \textbf{\SI{0.9885}{}} & \textbf{\SI{0.9924}{}} & \textbf{\SI{0.9849}{}} & \textbf{\SI{93.78}{\%}} & 2234 \\
\midrule

\multirow{3}{*}{\makecell[l]{DDPM + Transformer}}
& Action-Only Generator + Predicate-Only Predictor & 0.45 $\pm$ 0.10 & \SI{0.9364}{} & \SI{0.9355}{} & \SI{0.9561}{} & \SI{84.18}{\%} & 2800 \\
& Multi-task Baseline (Shared Encoder)  & 0.33 $\pm$ 0.08 & \SI{0.8684}{} & \SI{0.8705}{} & \SI{0.8679}{} & \SI{85.43}{\%} & 2800 \\
& \textbf{Predicate Action Skill (\approachname{})}             & \textbf{0.58 $\pm$ 0.15} & \textbf{\SI{0.9872}{}} & \textbf{\SI{0.9925}{}} & \textbf{\SI{0.9822}{}} & \textbf{\SI{91.98}{\%}} & 2606 \\
\midrule

\multirow{3}{*}{\makecell[l]{CFM + UNet}}
& Action-Only Generator + Predicate-Only Predictor & 0.31 $\pm$ 0.11 & \SI{0.8297}{} & \SI{0.7985}{} & \SI{0.8775}{} & \SI{82.11}{\%} & 2800 \\
& Multi-task Baseline (Shared Encoder)  & 0.45 $\pm$ 0.11 & \SI{0.9551}{} & \SI{0.9595}{} & \SI{0.9559}{} & \SI{84.88}{\%} & 2784 \\
& \textbf{Predicate Action Skill (\approachname{})}            & \textbf{0.56 $\pm$ 0.12} & \textbf{\SI{0.9819}{}} & \textbf{\SI{0.9945}{}} & \textbf{\SI{0.9712}{}} & \textbf{\SI{92.18}{\%}} & 2800 \\
\midrule

\multirow{3}{*}{\makecell[l]{CFM + Transformer}}
& Action-Only Generator + Predicate-Only Predictor & 0.62 $\pm$ 0.12 & \SI{0.8969}{} & \SI{0.9194}{} & \SI{0.8869}{} & \SI{75.15}{\%} & 2612 \\
& Multi-task Baseline (Shared Encoder)  & 0.45 $\pm$ 0.12 & \SI{0.8914}{} & \SI{0.9387}{} & \SI{0.8573}{} & \SI{81.43}{\%} & 2800 \\
& \textbf{\textbf{Predicate Action Skill (\approachname{})}}            & \textbf{0.72 $\pm$ 0.12} & \textbf{\SI{0.9813}{}} & \textbf{\SI{0.9932}{}} & \textbf{\SI{0.9706}{}} & \textbf{\SI{92.21}{\%}} & 2759 \\

\bottomrule
\bottomrule
\end{tabular}
}
\end{table*}

\begin{table*}[t]
\centering
\caption{\textbf{Single-policy rollout evaluation on RoboMimic (Kitchen\_d0 + Coffee\_d0).}
We evaluate \approachname{} and baselines on two RoboMimic manipulation tasks under realistic visual and geometric complexity, training a \emph{single} policy per task (no planning or composition). Results are averaged across Kitchen and Coffee Preparation, weighting each environment equally. Policy metrics report mean$\pm$SE over seven(7) seeded rollouts, with standard errors propagated across environments. Predicate classification metrics are macro-averaged across predicates within each environment, then averaged across environments.}
\label{tab:rollout-eval-robomimic_blocked}
\resizebox{\textwidth}{!}{
\begin{tabular}{@{}llcccccc@{}}
\toprule
\textbf{Setting} & \textbf{Approach} &
\multicolumn{1}{c}{\makecell{Policy}} &
\multicolumn{5}{c}{\makecell{Predicate Classification}} \\
\cmidrule(lr){3-3} \cmidrule(lr){4-8}
(Objective + Backbone)& &
Max Reward $\uparrow$ &
F1 $\uparrow$ & Recall $\uparrow$ & Precision $\uparrow$ & Accuracy $\uparrow$ & Observations \\
\midrule
\midrule
\multirow{2}{*}{\makecell[l]{DDPM + UNet}}
& Action-Only Generator + Predicate-Only Predictor& 0.93 $\pm$ 0.02 & \SI{0.9194}{} & \SI{0.9095}{} & \SI{0.9479}{} & \SI{83.98}{\%} & 10476 \\
& \textbf{Predicate Action Skill (\approachname{})}               & \textbf{0.96 $\pm$ 0.02} & \textbf{\SI{0.9340}{}} & \textbf{\SI{0.9271}{}} & \textbf{\SI{0.9537}{}} & \textbf{\SI{90.74}{\%}} & 10246 \\
\midrule

\multirow{2}{*}{\makecell[l]{DDPM + Transformer}}
& \textbf{Action-Only Generator + Predicate-Only Predictor} & \textbf{0.97 $\pm$ 0.02} & \textbf{\SI{0.9451}{}} & \textbf{\SI{0.9360}{}} & \textbf{\SI{0.9643}{}} & \textbf{\SI{88.36}{\%}} & 10087 \\
& Predicate Action Skill (\approachname{})              & 0.93 $\pm$ 0.04 & \SI{0.9185}{} & \SI{0.9050}{} & \SI{0.9531}{} & \SI{88.03}{\%} & 10103 \\
\midrule

\multirow{2}{*}{\makecell[l]{CFM + UNet}}
& \textbf{Action-Only Generator + Predicate-Only Predictor} & \textbf{0.99 $\pm$ 0.01} & \textbf{\SI{0.9467}{}} & \textbf{\SI{0.9407}{}} & \textbf{\SI{0.9574}{}} & \textbf{\SI{88.26}{\%}} & 10243 \\
& Predicate Action Skill (\approachname{})              & 0.91 $\pm$ 0.03 & \SI{0.8884}{} & \SI{0.8824}{} & \SI{0.9296}{} & \SI{82.47}{\%} & 10543 \\
\midrule

\multirow{2}{*}{\makecell[l]{CFM + Transformer}}
& Action-Only Generator + Predicate-Only Predictor& \textbf{0.97 $\pm$ 0.02} & \SI{0.9324}{} & \SI{0.9206}{} & \textbf{\SI{0.9612}{}} & \SI{86.07}{\%} & 9927 \\
& Predicate Action Skill (\approachname{})              & 0.89 $\pm$ 0.03 & \textbf{\SI{0.9386}{}} & \textbf{\SI{0.9287}{}} & \SI{0.9587}{} & \textbf{\SI{89.78}{\%}} & 10369 \\

\bottomrule
\bottomrule
\end{tabular}
}
\end{table*}

\section{Results and Discussion}
\label{sec:results}
\begin{figure*}   
  \centering
  \includegraphics[width=1\textwidth]{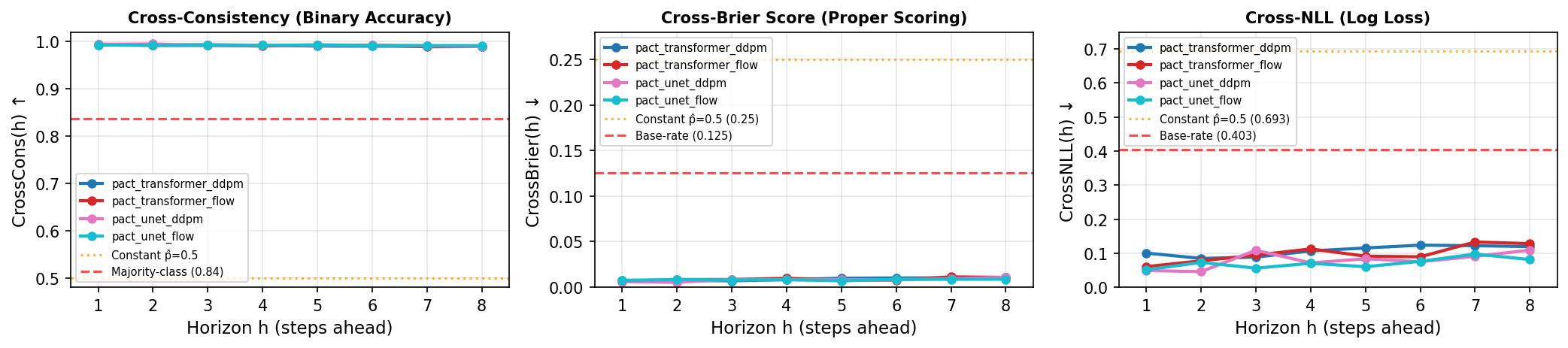}

\caption{\textbf{Predicate--action coherence of predicted belief rollouts.}
We measure how well \approachname{}'s predicted predicate-belief rollout matches the predicate outcomes realized after executing its open-loop action chunks.  The x-axis reports lookahead $h$ steps into the chunk. We plot (left) \textsc{CrossCons}$(h)$ (thresholded agreement; higher is better), (middle) \textsc{CrossBrier}$(h)$ (Brier score; lower is better), and (right) \textsc{CrossNLL}$(h)$ (log loss; lower is better). Horizontal lines are dataset-statistics baselines: constant $\hat p{=}0.5$ (dotted) and a base-rate/majority predictor from label frequencies (dashed). Since these baselines ignore both observations and planned actions, beating them indicates the beliefs reflect action-conditioned outcomes rather than priors. Across objectives and backbones, \approachname{} maintains high agreement and low losses over the horizon.}

\label{fig:coherence}
  \vspace*{-1mm}
\end{figure*}

\begin{figure*}   
  \centering
  \includegraphics[width=1\textwidth]{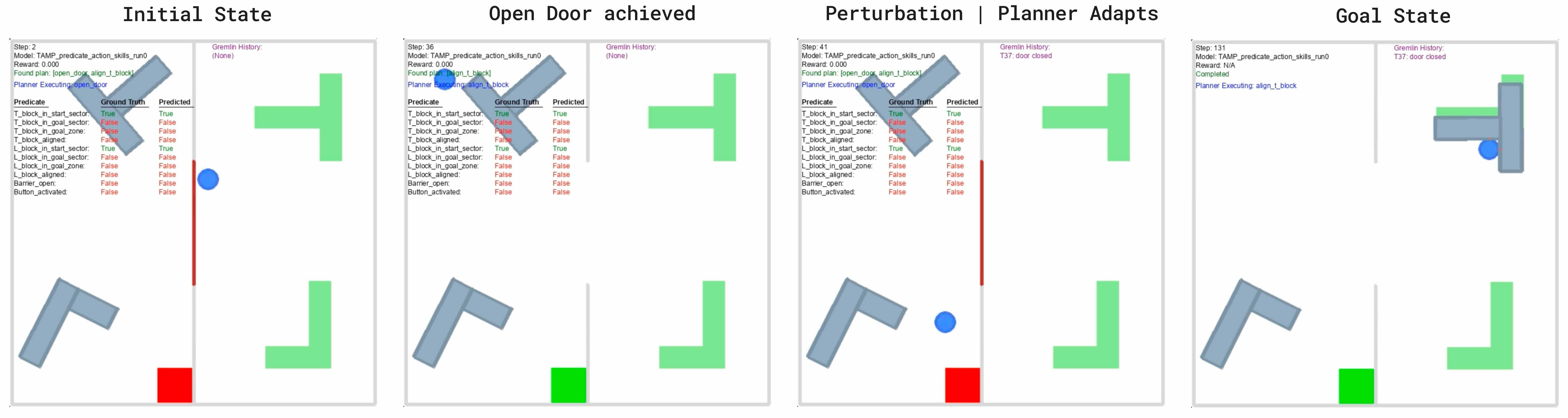}
\caption{\textbf{Skill recomposition and adversarial recovery in PushBarrier.}
Left to right: starting from the initial state, the planner synthesizes a sequence for a \emph{novel} subgoal (\texttt{open\_door} $\rightarrow$ \texttt{align\_t\_block}). After \texttt{open\_door} succeeds, an adversarial perturbation (\emph{Gremlin}) closes the door, which invalidates the current plan while aligning the $T$ block. Using online predicate-belief estimates to check effects, the system triggers replanning, re-invokes \texttt{open\_door}, and then resumes \texttt{align\_t\_block} to reach the goal.}

\label{fig:planning}
  \vspace*{-3mm}
\end{figure*}

\noindent\textbf{Single-policy analysis.}
We first study \approachname{} in the simplest setting: training a \emph{single} long-horizon policy per task. This isolates the impact of jointly generating action--outcome rollouts on policy performance and predicate prediction independent of planning, or composition.
Crucially, our goal is not to establish \approachname{} as the best monolithic policy learner, but to test whether we can \emph{add a planner-aligned outcome interface while preserving closed-loop control} and obtain an outcome trace that is (i) coherent under execution and (ii) suitable for downstream composition.

\paragraph{PushBarrier (2D compositional benchmark).}
Table~\ref{tab:rollout-eval-pushtlbarrier_blocked} reports single-policy performance and predicate classification on PushBarrier across objectives and backbones. Across all settings, \approachname{} achieves strong task reward while consistently improving predicate prediction and policy performance relative to baselines. Notably, the shared-encoder multi-task baseline often attains strong predicate scores, but its factorized heads yield weaker control; in contrast, \approachname{} samples \emph{paired} action--predicate rollouts from $p_\theta(\mathbf{x},\mathbf{z}\mid \mathbf{o}_t)$ and remains competitive (often best) on reward. This supports our primary modeling claim: adding outcome channels does \emph{not} require trading off control quality for symbolic prediction.
Instead, learning the coupled object $p_\theta(\mathbf{x},\mathbf{z}\mid \mathbf{o}_t)$ provides a useful inductive bias, encouraging the policy to represent \emph{which} action modes correspond to \emph{which} symbolic outcomes rather than learning actions and predicates as loosely related outputs.

Objective/backbone choices shift absolute numbers but not the conclusion: joint modeling remains strong across DDPM/CFM and UNet/Transformer. Consistent with prior work, UNet-style backbones are a reliable default, whereas Transformer variants can be more tuning-sensitive~\cite{chi2024diffusionpolicy}; in our evaluation, all models are trained for the same number of epochs without per-setting hyperparameter sweeps. We therefore view DDPM vs.\ CFM and UNet vs.\ Transformer primarily as implementation trade-offs (e.g., sampling speed and tuning sensitivity), with the primary differentiator being the \emph{joint} modeling target. \looseness=-1

\paragraph{RoboMimic (Kitchen\_d0 + Coffee\_d0).}
Table~\ref{tab:rollout-eval-robomimic_blocked} reports single-policy evaluation averaged across both RoboMimic tasks.
The action-only baseline remains strong on return across several objective/backbone choices, underscoring that explicit outcome modeling is not strictly necessary to achieve high reward when training end-to-end on a fixed task distribution.
Our objective, however, is planning-based composition, which requires a \emph{planner-aligned outcome interface} that stays coupled to the policy’s sampled actions and remains meaningful under execution---without sacrificing closed-loop control in realistic 3D manipulation.
In this setting, \approachname{} remains competitive in return and frequently improves predicate quality, with the strongest overall trade-off in the DDPM+UNet configuration.

\paragraph{Predicate--action coherence.}
An outcome interface is only useful for composition if it remains \emph{coupled} to what the policy actually does.
Per-step predicate metrics alone (Tables~\ref{tab:rollout-eval-pushtlbarrier_blocked}--\ref{tab:rollout-eval-robomimic_blocked}) do not test whether predicted outcomes match what the model’s \emph{own} generated actions induce.
Figure~\ref{fig:coherence} therefore evaluates predicate--action coherence: whether predicted belief rollouts agree with oracle predicate outcomes \emph{under the model’s executed actions} across lookahead horizons.
We report \textsc{CrossCons}$(h)$ and two proper scoring rules, \textsc{CrossBrier}$(h)$ and \textsc{CrossNLL}$(h)$, alongside constant and base-rate/majority reference baselines computed from dataset label statistics (i.e., priors).
Across objectives and backbones, \approachname{} achieves near-perfect \textsc{CrossCons} and substantially lower \textsc{CrossBrier}/\textsc{CrossNLL} than these reference baselines across the full horizon, indicating that predicted belief rollouts track outcomes realized under execution rather than reflecting label priors.
This provides empirical support for our central claim that $\mathbf{z}$ functions as an \emph{action-dependent outcome trace}, rather than an auxiliary prediction.

\noindent\textbf{Planning-based composition and recovery.}
We demonstrate planning-based recomposition of learned skills and replanning-based recovery under perturbations using the predicate-belief interface. Figure~\ref{fig:planning} shows a PushBarrier run with two key properties.
First, the goal is \emph{novel}: it specifies only a subgoal of the full demonstrated behavior (e.g., achieve \texttt{align\_t\_block} without completing the entire monolithic task).
Second, execution is perturbed adversarially: the door is closed mid-execution during \texttt{align\_t\_block}.
By checking symbolic effects using updated predicate beliefs, the system detects that the current plan is no longer applicable, triggers replanning, and recomposes a corrected sequence (\texttt{open\_door} $\rightarrow$ \texttt{align\_t\_block}) to reach the goal.

This illustrates two benefits of joint action--outcome skills:
(i) \emph{Recomposition}: skills expose planner-aligned predicates, allowing an off-the-shelf planner to synthesize sequences for goals not demonstrated end-to-end;
(ii) \emph{Recovery}: predicate beliefs make failures legible at the symbolic level, enabling replanning-based correction without retraining.

\section{Limitations}

\approachname{} inherits several limitations from predicate and skill learning.
First, performance depends on the choice and coverage of the predicate set: incomplete or mis-specified predicates that are misaligned with task structure can make planning brittle and limit the expressivity of the outcome interface.
Second, while joint modeling improves action--outcome coherence, it does not eliminate errors from perception and state aliasing---especially under occlusions and long horizons---and execution can still fail under stochasticity, partial observability, or distribution shift.
Finally, our composition results assume a manually defined symbolic domain (e.g., PDDL operators and goals); automatically discovering predicates, operators, or abstractions remains an open research direction.
A promising avenue is to leverage pretrained vision--language models to propose and evaluate candidate predicates directly from demonstrations \cite{athalye2026pixels}, then distill a compact planner-compatible set for downstream planning.

\section{Conclusion}
\label{sec:conclusion}

We introduced \approachname{}, a class of outcome-aware robot skills that jointly model action trajectories and predicate-belief trajectories by learning the coupled distribution $p_\theta(\mathbf{x},\mathbf{z}\mid\mathbf{o})$. Treating predicates as generated belief channels (rather than as a separate classifier output) yields \emph{paired} action--outcome rollouts that expose a planner-aligned interface while remaining compatible with modern trajectory generators. Overall, our results suggest that the right modeling object for composable skills is not action trajectories alone, but joint action--outcome rollouts whose symbolic trace is generated together with the policy's sampled action mode. We hope this perspective motivates future work on scaling predicate vocabularies and symbolic interfaces (e.g., via foundation-model priors) and learning richer abstractions that make planning-based composition reliable over longer horizons.

\footnotesize{

\bibliographystyle{IEEEtran} 
\bibliography{references}
}

\end{document}